# RankSum-An Unsupervised Extractive Text Summarization based on Rank Fusion

Akanksha Joshi[a,b,c,*], Eduardo Fidalgo [a,b] Enrique Alegre [a,b], Rocío Alaiz-Rodriguez [a,b]

[a]*Department of Electrical, Systems and Automation, Universidad de León, 25, 24004 León, Spain* [b]*Researcher at INCIBE (Spanish National Cybersecurity Institute), Av. de José Aguado, 41, 24005, León, Spain* [c]*Centre for Development of Advanced Computing, Mumbai, India*

∗*Corresponding author: ajos@unileon.es. Postal address: Centre for Development of Advanced Computing, Gulmohar Cross Road No.9, Juhu, Mumbai-400049, Maharashtra, India. Ph. 0091-8369817457*

*Email addresses: ajos@unileon.es(Akanksha Joshi), eduardo.fidalgo@unileon.es(Eduardo Fidalgo), enrique.alegre@unileon.es(Enrique Alegre), rocio.alaiz@unileon.es(Rocio Alaiz-Rodriguez)*

# RankSum-An Unsupervised Extractive Text Summarization based on Rank Fusion

Akanksha Joshi[a,b,c,*], Eduardo Fidalgo[a,b], Enrique Alegre[a,b],
Rocio Alaiz-Rodriguez[a,b]

[a]*Department of Electrical, Systems and Automation, Universidad de León, Spain*
[b]*Researcher at INCIBE (Spanish National Cybersecurity Institute), León, Spain*
[c]*Centre for Development of Advanced Computing, Mumbai, India*

**Abstract**

In this paper, we propose Ranksum, an approach for extractive text summarization of single documents based on the rank fusion of four multi-dimensional sentence features extracted for each sentence: topic information, semantic content, significant keywords, and position. The Ranksum obtains the sentence saliency rankings corresponding to each feature in an unsupervised way followed by the weighted fusion of the four scores to rank the sentences according to their significance. The scores are generated in completely unsupervised way, and a labeled document set is required to learn the fusion weights. Since we found that the fusion weights can generalize to other datasets, we consider the Ranksum as an unsupervised approach. To determine topic rank, we employ probabilistic topic models whereas semantic information is captured using sentence embeddings. To derive rankings using sentence embeddings, we utilize Siamese networks to produce abstractive sentence representation and then we formulate a novel strategy to arrange them in their order of importance. A graph-based strategy is applied to find the significant keywords and related sentence rankings in the document. We also formulate a sentence novelty measure based on bigrams, trigrams, and sentence embeddings to eliminate redundant sentences from the summary. The ranks of all the sentences -computed for each feature- are finally fused to get the final score for each sentence in the document. We evaluate our approach on publicly available summarization datasets- CNN/DailyMail and DUC 2002. Experimental results show that our approach outperforms other existing state-of-the-art summarization methods.

[*]ajos@unileon.es
*Email addresses:* ajos@unileon.es (Akanksha Joshi),
eduardo.fidalgo@unileon.es (Eduardo Fidalgo), enrique.alegre@unileon.es
(Enrique Alegre), rocio.alaiz@unileon.es (Rocio Alaiz-Rodriguez)

## 1. Introduction

In the information age, a massive amount of textual data is available, and text summarization plays a critical role in getting quickly key information within the text content. The objective of text summarization is to generate a condensed representation of a document while preserving the significant content. Text summarization is widely applicable to many areas such as news summarization, social media (blogs or tweets) summarization, event summarization, email summarization, review summarization, legal documents summarization, scientific articles summarization, sentence compression, and question answering, among others (El-Kassas et al., 2021). However, manual text summarization consumes a lot of time, effort, economic resources and becomes unfeasible for many such tasks (El-Kassas et al., 2021).

Text summarization can be broadly classified as extractive (Mihalcea and Tarau, 2004; Cao et al., 2015; Ren et al., 2017; Nallapati et al., 2017a; Cheng and Lapata, 2016; Zhou et al., 2020; Joshi et al., 2019; Liu, 2019) or abstractive type summarization (Rush et al., 2015; Nallapati et al., 2016; Wang and Ling, 2016; Rush et al., 2015; El-Kassas et al., 2021; Gambhir and Gupta, 2017). Extractive text summarization generates summaries selecting the significant sentences in the document whereas, abstractive summarization produces summaries by paraphrasing the relevant content in the document. Abstractive may seem to be the best way to generate a summary, but solutions are still preliminary, although promising results have been found with deep learning approaches (Zhang et al., 2019a; Chopra et al., 2016). Moreover, abstractive summaries suffer from the problem of word repetition and out-of-vocabulary words (El-Kassas et al., 2021).

The extractive summarization approach (Erkan and Radev, 2004; Cheng and Lapata, 2016; Baralis et al., 2013) is faster and simpler than the abstractive one. It also leads to higher accuracy (Tandel et al., 2016) because the summary uses the exact terminologies from the original document. Nonetheless, it is far from the way human experts write summaries (Hou et al., 2018) because there may be redundancy in some summary sentences and it might lack semantics and cohesion in summary sentences (Moratanch and Chitrakala, 2017).

Extractive summarization can be further classified as supervised or unsupervised. Supervised methods (Nallapati et al., 2017a; Cheng and Lapata, 2016) require training data and also to model text summarization as a binary classification problem that tags the sentences as summary or non-summary. One of the main drawbacks of supervised learning approaches (Ren et al., 2017; Nallapati et al., 2017a; Cheng and Lapata, 2016; Zhou et al., 2020; Joshi et al., 2019; Liu, 2019; Bahdanau et al., 2015) is that they require a lot of labeled training data to produce good extractive summaries. The unsupervised text summarization methods do not require labeling, and ranking the sentences depending on various sentence features such as keywords, term frequency, sentence position, sentence length, and other factors (Erkan and Radev, 2004; Filatova and Hatzivassiloglou, 2004). A score is assigned to the sentences according to these features, and finally the sentences are chosen using graph-based methods (Erkan and Radev, 2004; Mihalcea and Tarau, 2004; Parveen et al., 2015), greedy approaches (Carbonell and Goldstein, 1998), or optimization-based summarization techniques (McDonald, 2007).

Several extractive methods have been proposed in the literature (Gambhir and Gupta, 2017), and each has its own advantages, and limitations. In this paper, we propose to combine different techniques to benefit from their specific advantages and reduce their limitations, which should enable the generation of better summaries.

Our proposal is a unified framework for extractive text summarization that is fully unsupervised and merges the ranks obtained by different techniques. It relies on methods based on topic, keywords, semantics, and positional information and unlike other fusion strategies, the merging occurs at the rank level. It is easy to fuse at rank level rather score level due to incompatibility and normalization issues present at score level due to scores obtained via different methods.

The topic content (Blei, 2012) captures the global saliency of a document and has been implemented for understanding long-range dependencies inside documents (Mikolov and Zweig, 2012). The sentence embeddings preserve the semantic meaning of the sentences. We use siamese networks (Bromley et al., 1993) with triplet loss to derive sentence embeddings for our task. These embeddings efficiently represent the semantics of sentences and can be efficiently utilized for summarization tasks. Several approaches have been applied in literature (Jindal and Kaur, 2020; Litvak and Last, 2008; Matsuo and Ishizuka, 2003) to derive keywords in the text for summarization purposes. It is based on the assumption that significant sentences contain the significant keywords of the document. The other attribute that we employed in our approach is relative positioning of a sentence in the document. Additionally, to identify redundancy in the summary text, we use sentence embedding, bigrams, and trigrams. Through experiments, we showed that each sentence feature is significant for generation of good summaries, however, different features complement each other and can produce a more meaningful representation of the document. Our main contributions are summarized as follows:

1. We generate a novel topic rank for each sentence based on probabilistic topic models. The topic score of each sentence is computed by estimating the distance of topic representation of each sentence from the topic centroid of the document. The significant sentences in the document fall close to the topic centroid of the document.

2. We introduce a new method for ranking sentences based on sentence semantic embeddings that can efficiently capture the meaning of each sentence in the document. We recursively determine document embedding by removing each sentence from the document and calculate the difference each time with the document embedding computed using all the sentences of the document.

3. We also formulate a novelty parameter based on bigrams, trigrams, and sentence embeddings to eliminate the redundant sentences from the summary.

4. We propose RankSum, a unified framework for extractive text summarization that summarizes documents based on multi-dimensional sentences features - topic information, semantic content, keywords, and sentence position- in the document. RankSum ranks the sentences of documents based on each of these features and then finally performs a weighted rank level fusion to generate a final summary.

5. Finally, we evaluated our summarization method on publicly available summarization datasets- DUC 2002 and CNN/DailyMail. Empirically, we demonstrated that our unsupervised summarization approach is quite robust as compared to other state-of-the-art proposals, including the supervised methods, on both datasets.

The paper is structured as follows: Section 2 presents the related literature in the domain. Section 3 discusses the proposed summarization strategy whereas Section 4 evaluates the approach on publicly available summarization datasets. Section 5 provides our main conclusion and future work.

## 2. Related work

The extractive text summarization approach has been applied using many different methods. We can broadly classify them as statistical-based, concept-based, optimization-based (Sanchez-Gomez et al., 2020), topic-based, graph-based, sentence centrality-based, semantic-based, deep learning-based (Cheng and Lapata, 2016). Statistical-based methods (Gupta and Lehal, 2010) select important sentences and words for summary depending on their position and most frequent terms or keywords in the sentence. Concept-based summarization (Moratanch and Chitrakala, 2017) includes retrieving the concepts from an external knowledge source, building a graph model to find relations between concepts and sentences and then applying a ranking algorithm to score sentences. Sentence Centrality-based methods (Erkan and Radev, 2004) extract the most important and central sentence in a cluster using the centrality of words which is estimated using the centroid of a document cluster. Topic-based approaches (Mihalcea and Tarau, 2004) focus on identifying significant sentences based on the topics of the document, which are estimated using term frequency, term frequency-inverse document frequency, or lexical chains, among others. Graph-based methods build a graph of the document to identify the relationships among sentences and then use a ranking algorithm to determine summary sentences. Semantic-based methods (Mohamed and Oussalah, 2019) identify key sentences through methods that explore text semantics such as Latent Semantic Analysis (LSA), Semantic Role Labeling (SRL), and Explicit Semantic Analysis (ESA). Optimization-based methods utilize an optimization algorithm such as sub-modular programming, Multi-Object Artificial Bee Colony Algorithm to generate a summary of length $L$. Deep Learning methods (Cheng and Lapata, 2016) mainly aim at applying deep neural networks such as Convolution Neural Networks (CNN), Recurrent Neural Networks, and their variants to achieve automatic text summarization.

A recent survey on text summarization (El-Kassas et al., 2021) illustrates the advantages and drawbacks of each one of the approaches mentioned above. However, by combining different methods in a right way, we may achieve better summaries because we benefit from each method's advantages. This paper proposes a method that combines several approaches such as topic, graph (keyword), semantic, and statistical-based (position). Afterward, we will review the related approaches one by one.

Several proposals have been presented for keyword-based summarization. Litvak and Last (2008) introduced a supervised and unsupervised graph-based approach for keyword extraction to summarize documents. They used the HITS algorithm to find keywords and showed that supervised classification gives the highest keyword identification accuracy given a large labeled training set. Their experiments showed that the supervised approach works better in the case of enough data, whereas the unsupervised method provides higher accuracy in case of unavailability of labeled data for training. Fattah and Ren (2009) took into account several features, including sentence position, positive keyword, negative keyword, sentence centrality, sentence resemblance to the title, sentence inclusion of name entity, sentence inclusion of numerical data, relative sentence length, the bushy path of the sentence then trained a summarizer using all of them and analyzed that some features like keywords are language-dependent and some are language independent. Their keyword-based summarization lags behind various other textual features such as bushy path, sentence centrality, sentence length, sentence resemblance to title, and position. Mihalcea and Tarau (2004) introduced the graph-based ranking model from natural language texts for unsupervised keyword and sentence extraction tasks that achieved state-of-the-art accuracies. They illustrated that their keyword extraction approach lagged in precision behind the supervised method. Baralis et al. (2013) presented a novel graph-based summarization approach that exploits association rules among various terms in a document for summarization. Ferreira et al. (2014) also introduced a graph-based approach that converts text into a graph model consisting of four types of relations between sentences: similarity statistics, semantic similarity, co-reference, and discourse relations. Their approach lacks sentence ordering as their system finds sentences in a group of different topics for summarization.

Gialitsis et al. (2019) examined the effects of probabilistic topic model-based word representations for extractive text summarization based on supervised algorithms such as Naive Bayes, Quadratic Discriminant Analysis, or Gradient Boosting Classifiers. They demonstrated that topic modeling outperforms TF-IDF for sentence classification for extractive summarization tasks. Gao et al. (2012) applied LDA to identify semantic topics in the document and then construct a bipartite graph to represent the document and further find sentence salience scores of sentences and topics simultaneously. Hennig (2009) developed a method for query-focused multi-document summarization based on Latent Semantic Analysis (LSA) to represent sentences and queries as probability distributions over latent topics. Their analysis proved that LSA provides better results as compared to Latent Semantic Indexing (LSI). Ailem et al. (2019) explored the use of latent topic information to reveal more global content, which can be used to bias the decoder network to generate words for abstractive summarization. Nagwani (2015) designed a technique using semantic similarity-based topic modeling and topic models to summarize documents over the MapReduce framework. They showed that for effective summarization, we need both semantic information and clustering. Recently, Narayan et al. (2018a) introduced an extreme summarization task for news articles that use topic information in a document and Convolution Neural Networks(CNN). Our method is quite different from the approaches mentioned above in how we use topic vectors to arrange sentences according to their respective salience. We exploit the document topic vector obtained using LDA to determine how much a sentence is close or far from the document centroid. Our method is different from Parveen et al. (2015) in how they are utilizing the topic information obtained using LDA. They constructed a graph between sentences and topics in the document and then applied the HITS (Kleinberg, 1999) algorithm to rank sentences. They only used word topic vectors to assign weights to edges in the graph but we used both word topic vector and document vector in our system. Moreover, we are trying to estimate how each sentence is closer to the topic vector of the document to find the importance of each sentence.

To use sentence embeddings, Bouscarrat et al. (2019) introduced STRASS. This extractive text summarization method creates a summary by leveraging the semantic information in existing sentence embeddings spaces in both supervised and unsupervised fashion. Their approach failed to handle the case of multiple topics summaries and thus missed some topics in summaries. Their embedding model is fast and light to run on a CPU. Liu and Lapata (2019) presented a novel document-level encoder based on embeddings generated using BERT architecture which can express the semantics of a document and obtain representations for sentences. They showed that their BERT-based document representations can attain state-of-the-art accuracy for both extractive and abstractive summarization. Some other supervised deep learning methods (Cheng and Lapata, 2016; Nallapati et al., 2017a,b; Zhou et al., 2018; Liu, 2019; Narayan et al., 2018c; Tarnpradab et al., 2017; Zhang et al., 2018; Wu and Hu, 2018; Narayan et al., 2017) that implicitly derive summarization specific sentence representations have been applied for extractive summarization are known to have state-of-the-art accuracies. Joshi et al. (2019) proposed SummCoder, an unsupervised approach based on deep auto-encoders that exploited skip-thoughts sentence vectors to generate extractive summaries. Their approach also took advantage of multiple sentence features such as position, novelty, and sentence saliency computed using auto-encoders. None of the proposed methods used siamese networks for producing sentence embeddings, and we formulated a new unsupervised method to rank sentences using them. Our sentence ranking algorithm differs from Joshi et al. (2019) in the manner as authors generate document embeddings recursively using auto-encoders, whereas we produce document embeddings via averaging sentence vectors.

Lamsiyah et al. (2020) used pre-trained sentence embeddings as an input to Feed Forward Neural Networks (FFNN), a supervised method. However, in our case, we fine-tuned the SBERT network on our dataset to derive optimal embeddings and then proposed a novel unsupervised strategy to identify the ranks of sentences using the derived embeddings. The Centroid-based text summarization method presented by Rossiello et al. (2017) exploited word embeddings rather than sentence embeddings and calculated centroid by summing the vectors of important words in the document identified using term frequency-inverse document frequency vector. The sentence is ranked by simply estimating the distance of the document centroid with sentence embedding, which is obtained by summing word embeddings. Our Semantic Rank Extractor differs from this method because we employed sentence embeddings that better represent the meaning of a sentence compared to word embeddings. Our centroid calculation and sentence saliency estimation strategy are distinct from them. We obtained document embeddings by summing sentence embeddings rather than word embeddings. The sentence ranking method iteratively estimates document centroid to rank each sentence in the document. Ramirez-Orta and Milios (2021) employed a graph-based method over sentence embeddings to find the rank of each of the sentences whereas, we proposed an altogether different strategy in which document centroid is computed iteratively by removing one sentence at a time from the document and then estimating the difference from the original document centroid to obtain each sentence ranking.

Few summarization approaches have explored the fusion of different summarization features or algorithms. Dutta et al. (2018) presented an ensemble algorithm that combined the outputs of multiple summarization algorithm to produce final summaries. They developed an unsupervised graph-based strategy and a supervised method Learning-to-Rank to fuse the output of multiple algorithms. The Summcoder approach of Joshi et al. (2019) introduced a method based on the weighted fusion of three sentence features-relevance, novelty and position. You et al. (2020) presented topic information fusion and semantic relevance for text summarization based on Fine-tuning BERT(TIF-SR). Firstly, the authors extracted topic keywords and fused them with source documents. Then they made the summary closer to the source document by computing the semantic similarity between the generated summary and the source document. This approach requires labeled data for summarization. Wong et al. (2008) designed a learning-based approach using various sentence features such as surface, content, relevance, and event. They combined all the features using semi-supervised learning to minimize the dependency on labeled datasets for summarization. However, their approach needs labeled data and thus, dependent on the domain for which it is trained. Mao et al. (2019) developed three methods to fuse and score sentences by combining sentence relations with statistical features of sentences using supervised and unsupervised learning. This method only explored statistical features and sentence relations with each other and would have missed other features required to build an optimal summarization system. Palshikar et al. (2012) selected a few summarization approaches from literature and averaged the ranks obtained via different summarization methods proposed in the literature to get the summary. In our paper, we have given weights to ranks assigned via different ranking techniques rather averaging them. In another summarization system, Barrera and Verma (2012) performed score fusion based on the syntactic and semantic features of the sentences. However, in our proposal, we performed a rank fusion exploring the topic, semantic, keyword, and positional information. Meena and Gopalani (2014) used several combinations of various sentence features such as Tf-idf, word co-occurrence, sentence centrality, among others, for summarization. Still, they missed semantic and topic information in the document, which is crucial to producing a meaningful summary (Parveen et al., 2015). Moreover, no weighted fusion in the paper has been explored, which might have yielded a better accuracy.

Our approach is quite distinct from all of the above approaches in the manner that we propose the fusion of all the sentence features based on topic, sentence embeddings, keywords, and position. Finally, we combine all of these features using weighted rank fusion that is not proposed by any of the above algorithms. Moreover, our summarization method is completely unsupervised, which is different from other fusion strategies that used supervised and unsupervised learning and required labeled training data.

## 3. Proposed RankSum Approach

### 3.1. Problem formulation

Let each document $D$ consist of $N$ sentences ($S_1$, $S_2$, ..., $S_N$) and $M$ words as ($w_1$, $w_2$, ..., $w_M$). The goal of a summarization framework is to extract a ranked set of the top-L most significant sentences from the document to represent it in a compressed form, i.e. a summary of $L$ sentences. The summarization approach proposed, which we named RankSum, uses four multi-dimensional sentence features (topics, keywords, semantics, and position) to rank sentences in a document. First, we generate a rank for each sentence in the document according to each feature, gathering information from different aspects of the sentences. Then, we compute a weighted rank fusion derived from the four generated ranks. We hypothesize that each sentence feature contributes to generate a good summary, and we assume that different features complement each other and can produce a more meaningful representation of a document.

The overall pipeline of the proposed RankSum framework is given in Figure 1. In the sections below, we discuss each of the features used to rank

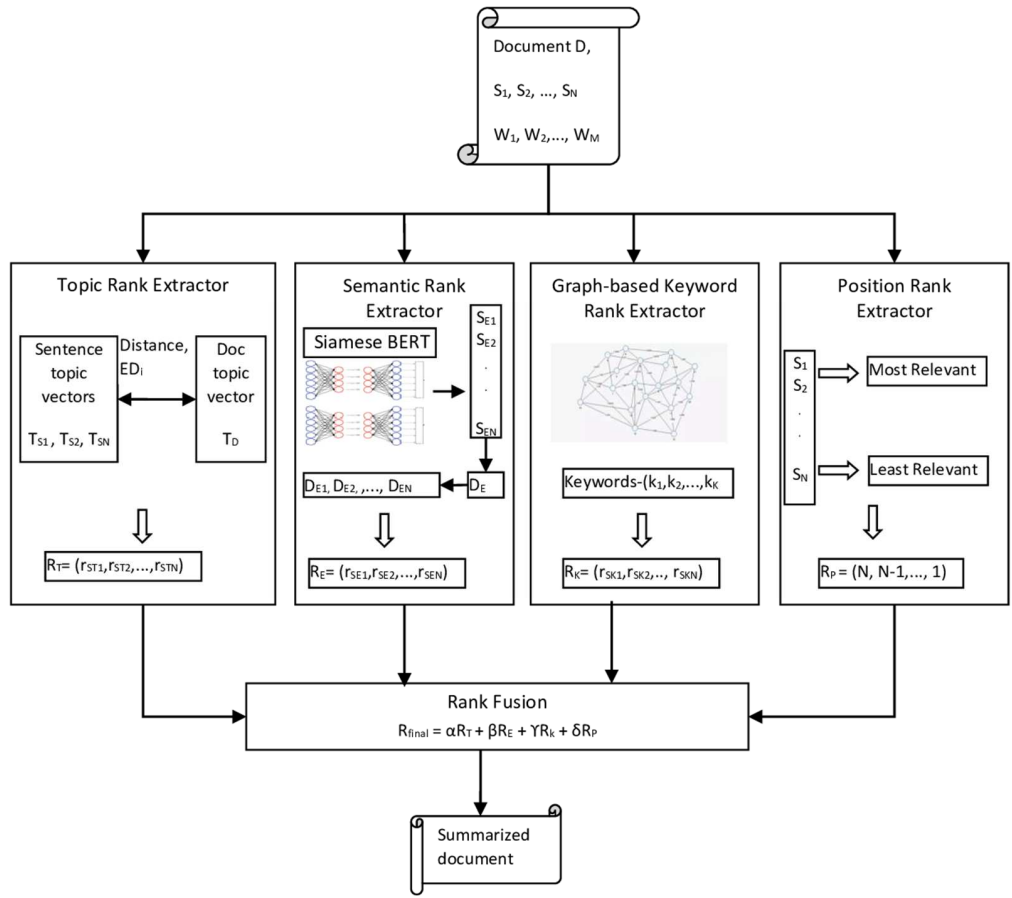

Figure 1: Schema of the RankSum architecture

sentences and the fusion methodology applied, in more detail.

*3.2. Topic Rank Extractor*

In this section, we present a new method to rank sentences based on their topic vectors. We assume that topic vectors provide a latent representation of documents which is quite significant while summarising documents. The topic information has been used in query-oriented (Hennig, 2009), abstractive (Ailem et al., 2019) and multi-document summarization (Nagwani, 2015) to boost the summarization accuracy and to complement their method with additional information. Topic information can preserve the global meaning of a document, which is helpful in summarization to understand the long-range semantic information in the text. We employ Latent Dirichlet Allocation (LDA)(Blei et al., 2003), to find the topics in the document. LDA models a text document as a mixture of topics and each topic as a collection of words that tend to co-occur together. Each topic is represented as a probability distribution of key terms in the text document and each document is modeled as a probability distribution of topics. Thus, LDA computes the topic-term distribution and document-topic distribution from a large collection of documents using Dirichlet priors for distributions over a fixed number of topics. The LDA consists of the following steps:

1. Firstly, scan each document and randomly assign each word to one of the K topic vectors
2. For each Document d, go through each word w and compute:
    (a) *probability*(*topic, t | document, d*), the words in document d that are assigned to topic t
    (b) *probability*(*word, w | topic, t*), captures the proportion of documents in topic t because of word w. If a word is having a high

probability of belonging to topic t, then all the documents having word w have a high probability of being associated with topic t.

(c) Reassign w a new topic, t as:

$$probability(w,t) = probability(t|d) * probability(w|t) \quad (1)$$

By applying LDA, we first calculate the topic vectors $T_D$ for each document in the corpus and topic vectors, $T_w$ for each word in the document. Then we obtain the sentence topic vector, $T_{Si}$ by averaging the topic vectors of each word present in the sentence. To rank the sentences in the document, we compute the euclidean distance, $ED_i(T_{S_i}, T_D)$ between the topic vector of each sentence, $T_{S_i}$ and the topic vector of the document $T_D$ as given in Eq. 2

$$ED_i(T_{S_i}, T_D) = \sqrt{\sum_{q=1}^{Q}(T_{S_{iq}} - T_{D_q})^2} \quad (2)$$

where Q is the length topic vector for sentence, $i$ and document, $D$. The more important sentences will fall close to the document topic vector and ranked accordingly. We represent the rank topic vector generated for a document as

$$\mathbf{R_T} = (r_{ST_1}, r_{ST_2}, \ldots, r_{ST_N}) \quad (3)$$

where $r_{ST_i}$ is the topic rank associated with each sentence $i$, $1 \leq r_{ST_i} \leq N$ and $N$ is the total number of sentences. Note that 1 is considered the lowest rank.

*3.3. Embedding-based Semantic Rank Extractor*

In order to identify significant sentences based on their semantics, we exploit sentence embeddings. We use SBERT (Reimers and Gurevych, 2019) to obtain the sentence embeddings for each sentence in the document. SBERT is a BERT-based architecture that utilizes Siamese and triplet networks to derive semantically meaningful embeddings. SBERT has been shown to outperform other state-of-the-art embeddings (Devlin et al., 2019; Conneau et al., 2017; Liu et al., 2019) on 7 Semantic Textual Similarity (STS) tasks. SBERT is also computationally efficient compared to other sentence embeddings.

We develop a novel algorithm to find the ranking of sentences based on their respective embeddings. Let $S_{E_i}$, represent the embeddings obtained using SBERT architecture for each sentence, $S_i$ in the document. We calculate the document embedding $D_E$ by averaging the embeddings of all the sentences in the document. To identify the saliency of each sentence in the document, we remove that sentence from the document and again obtain a new document embedding $D_{E_i}$. Next, to measure the saliency of the sentence $S_i$ in the document, we calculate the euclidean distance, $d_E$ between document vectors, $D_E$ and $D_{E_i}$. The notable sentence will generate high value of $d_{E_i}$ as compared to the sentences which do not express the meaning of document. Thus, we produce the rank vector, $\mathbf{R_E}$ for all sentences of the document based on their $d_{E_i}$ scores.

*3.4. Keyword Rank Extractor*

Keywords capture the structural content information in a document. The sentences that have keywords in them carry significant information compared to other sentences. To compute the set of keywords, $K = (k_1, k_2, ...k_K)$ in a document, we first remove the stop words and apply lemmatization. Then, we follow a graph-based strategy (Brin and Page, 1998) to identify the key terms in a document.

We generate the rank $\mathbf{R_K}$ regarding the keywords within each sentence in the document. We choose to give a higher rank to the sentences that consist of more keywords. If some sentences contain the same number of keywords, we rank them according to their positions. We assume that important sentences contain more keywords as compared to sentences that have less number of important words.

*3.5. Position Rank Extractor*

The relative position of a sentence in a document indicates the importance of sentence for summary generation (Luhn, 1958; Edmundson, 1969) and therefore, we use it as one relevant attribute for ranking sentences in the document. The sentences that appear at the beginning of a document are more relevant compared to the sentences which appear later in the document (Gupta et al., 2011). We generate the position rank vector, $\mathbf{R_P}$ by assigning a rank to sentences depending on their position. The sentence in the first position is given the highest rank and the last sentence is given the lowest rank as given in Eq. 4.

$$\mathbf{R_p} = N, N-1, .., 1 \qquad (4)$$

where, N is the total number of sentences in the document

*3.6. Sentence Novelty Extractor*

To eliminate redundant sentences while producing extractive summaries of the document, we propose a new sentence novelty extractor that makes use of sentence representations $S_{E_i}$ as obtained in section 3.3 as well as bigrams and trigrams present in the sentences. By finding the number of bigrams and trigrams, we can predict which two sentences are similar to each other, however, it ignores the semantics of the sentences. To overcome this issue, we complement our novelty extractor with sentence embeddings that are quite good at finding out semantically similar sentences. Embeddings generated using the SBERT network (Reimers and Gurevych, 2019) are quite robust and do well while predicting similar sentences during summary generation. We estimate the sentence novelty, $S_{Nov_i}$ as given in Eq. 5

$$S_{Nov_i} = \begin{cases} 1, & Sim(S_{E_i}, S_{E_j}) < t1 \quad or \quad Count_{bigrams,trigrams}(i,j)) < t2, \\ & 1 \leq j \leq V, \quad i \neq j \\ 0, & otherwise \end{cases} \quad (5)$$

where 1 indicates the sentence is novel and 0 tells that the sentence is redundant. V is the number of sentences that have been already added to the summary, t1 and t2 are the thresholds set experimentally to identify similar sentences, $Count_{bigrams,trigrams}(i,j)$ is the number of bigrams and trigrams that match between sentence $S_i$, and $S_j$ and $Sim(S_i, S_j)$ is the cosine distance between sentence $S_i$ and $S_j$ given by

$$Sim(S_i, S_j) = \frac{\vec{S_i}, \vec{S_j}}{\parallel \vec{S_i} \parallel \parallel \vec{S_j} \parallel} \quad (6)$$

*3.7. Rank fusion and summary generation*

We finally combine the information provided by the aforementioned modules at the rank level. We fuse all the rank vectors, $\mathbf{R_T}$, $\mathbf{R_K}$, $\mathbf{R_E}$, $\mathbf{R_P}$ and generate a final rank for each sentence in the document. The rankings are merged as given in Eq. 7

$$\mathbf{Rank_{final}} = \alpha \cdot \mathbf{R_T} + \beta \cdot \mathbf{R_K} + \gamma \cdot \mathbf{R_E} + \delta \cdot \mathbf{R_P} \quad (7)$$

where, values of *α, β, γ,* and *δ* are determined empirically. The final rank **Rank$_{final}$** determines the order in which the sentences will be added to the summary. Following an iterative process, a new sentence is added to the summary if it is distinct from the already added summary sentences based on the novelty extractor, $S_{Nov_i}$ given by Eq. (5) and defined in section 3.6.

**4. Experimental results and analysis**

*4.1. Datasets*

We used CNN/DailyMail dataset (Hermann et al., 2015) for training the siamese network for sentence embeddings. This dataset comprises 197, 000, and 90, 000 news articles from CNN and DailyMail which are used frequently in question-answering tasks. We divided the documents in CNN/DailyMail into training, validation, and testing as indicated in Table 1. We used the CNN/DailyMail test set and DUC 2002[1] dataset for evaluating our proposed approach and other algorithms for extractive text summarization. DUC 2002 is a standard summarization benchmark that consists of 567 news articles from 59 categories with at least two gold summaries for each of the articles.

*4.2. Experimental set up*

Firstly, we split the document into sentences and tokenize it into words. We removed the stop words and applied lemmatization for keyword rank generation. The topic vectors have been determined by applying LDA on the training set of CNN/DailyMail as mentioned in Table 1. We kept the length of the topic vector to 512 dimensions. The values of *α, β, γ,* and *δ* are set to 0.3, 0.35, 0.34, and .01 in Eq. 7, which are determined empirically using CNN/DailyMail Validation set. We assigned a weight of 0.01 to the position parameter because we determined empirically that if we kept it higher, it decreased the contribution from topic, keywords, and embeddings decreasing the overall accuracy. Although the contribution of the position parameter is smaller than the others, we also checked empirically that if we remove it, the accuracy of RankSum decreases. Finally, we are aware that sentence position may not be an effective indicator of importance in non-news text documents and also by keeping its weight value low, we can easily adapt our proposed framework to summarize the non-news datasets.

---
[1]https://duc.nist.gov/data.html

Table 1: Datasets used for training and evaluation.

| Dataset | Type | Usage | # Documents | # Categories |
|---|---|---|---|---|
| **CNN/DailyMail** | News | Training | 287,227 | - |
| | | Validation | 13,368 | - |
| | | Testing | 11,490 | - |
| **DUC 2002** | News | Testing | 567 | 59 |

*4.3. Evaluation*

To evaluate RankSum, we used ROUGE-1, ROUGE-2, and ROUGE-L metrics (Lin, 2004). ROUGE-1 and ROUGE-2 measure the unigram and bigram matches between the candidate and gold summary, whereas ROUGE-L gives the longest common subsequence matches between the candidate and gold summary.

We compared RankSum with other widely known state-of-the-art approaches for automatic summarization. We used a summary length of 100

words for DUC 2002 and a full-length ROUGE metric for CNN/DailyMail dataset. We used stemming and chose the best ROUGE-N scores available for two or more reference summaries available in DUC 2002. We picked the results for comparison directly from the papers. Not all the approaches reported accuracies on both datasets. Therefore, we illustrate the results accordingly on CNN/DailyMail and DUC 2002 datasets.

The following techniques are used for comparison with both datasets.

**LEAD** selects the first three leading sentences from the document to generate a summary.

**NN-SE** (Cheng and Lapata, 2016) is a method to jointly score and select sentences using the hierarchical encoder-decoder network.

**SummaRuNNEr** (Nallapati et al., 2017a) is an extractive summarization method based on Recurrent Neural Networks.

**HSSAS** is a hierarchically structured encoder-decoder network for self-attention proposed by Al-Sabahi et al. (2018).

**Rank-emb** is an extractive summarization method where we calculated the ranks of sentences using our semantic rank extractor.

**Rank-topic** summarizations strategy is based on deriving the ranks of sentences using our topic rank extractor.

**Rank-keyword** method exploits our keyword rank extractor to generate extractive summaries of the documents.

Additionally, we reported the accuracy on the CNN/DailyMail dataset for the following methods:

**Bi-AES** (Feng et al., 2018) used a bi-directional encoder with attention to find extractive summaries of the documents.

**REFERESH** (Narayan et al., 2018b) is another reinforcement learning-based approach that globally optimizes ROUGE evaluation metrics.

**PACSUM(BERT)** (Zheng and Lapata, 2019) is an unsupervised summarization algorithm that employed BERT to capture sentence similarity. It builds graphs with directed edges, arguing the contribution of any two nodes to their respective centrality is influenced by their relative position in the document.

**RNES**, developed by Wu and Hu (2018), is a reinforced learning-based extractive summarization for producing coherent summaries.

**JECS** (Xu and Durrett, 2019) consists of a sentence extraction model joined with a compression classifier that decides whether or not to delete syntax-derived compression for each sentence.

**NeuSum** presented by Zhou et al. (2020) is a neural network framework to score and select sentences for summary generation jointly.

**HIBERTM** (Zhang et al., 2019b) is Hierarchical Bidirectional Encoder Representations from Transformers for document encoding and a method to pre-train using unlabeled data for text summarization.

**BERTSum** proposed by Liu (2019) fine-tuned BERT architecture for extracting meaningful summaries from the document.

**BERTSUM+Classifier** (Liu, 2019) is a simple classifier developed for extractive summarization based on BERT architecture with inter-sentence transformer layers.

**BART** (Lewis et al., 2020) is a denoising auto-encoder built using sequence to sequence models that can be applied to a wide variety of tasks including summarization.

**PEGASUSLARGE** (Zhang et al., 2019a) is a large transformer-based encoder-decoder model pre-trained on massive text corpora where meaningful sentences are removed/masked from an input document and are generated together as one output sequence from the remaining sentences, similar to an extractive summary.

**MATCHSUM** (Zhong et al., 2020) is a summary-level framework that conceptualized extractive summarization as a semantic text matching problem.

For the DUC 2002 dataset we report the performance of the following approaches:

**Integer Linear Programming (ILP)** proposed by Woodsend and Lapata (2012) an ILP formulation to efficiently search through Quasi-synchronous grammar rules to provide a globally optimal solution for coherent and grammatical summaries.

**Tgraph** (Parveen et al., 2015) is a graph-based approach that uses topical information to compress the document with relevant information.

**URANK** is a unified ranking methodology presented by Woodsend and Lapata (2012) to simultaneously summarize both single and multiple documents.

**SummCoder** (Joshi et al., 2019) is an unsupervised auto-encoder based approach to find extractive summaries of the documents.

**CoRank** is proposed by Fang et al. (2017) that explores the word-sentence relationship for unsupervised summary extraction.

*4.4. Results*

Table 2 shows the results of our proposed RankSum framework and other state-of-the-art on the DUC 2002 dataset using ROUGE metrics. Our proposal achieves ROUGE-1, ROUGE-2, and ROUGE-L scores of 53.2, 27.9, and 49.3, respectively, outperforming all the recent methods analyzed for the extractive text summarization dataset. We exceeded the highly accurate summarization system, HSSAS, and Co-Rank, with a very high margin of 0.6, 0.8, and 0.5 for ROUGE-1, ROUGE-2, and ROUGE-L scores. Our approach is unsupervised, it does not require any labeled data to train the system, which is the requirement of most recently proposed summarization methods such as SummaRuNNer, HSSAS, and NN-SE. The proposed RankSum summarization method covers every critical aspect of summarization. Rather than focusing on just one feature, we make use of different features of sentences. It can also be observed that RankSum approach performs better when compared to individual rank extractors: Rank-emb, Rank-topic, and Rank-keyword. The topic vectors provide the global content of the document whereas keywords can capture the local structural information. The semantics of the sentences are well captured using sentence embeddings. The RankSum summarization strategy can supersede the results of other state-of-the-art methods, even supervised ones. The other methods do not consider the topic content in the document and, since we are using the improved SBERT embedding to represent the sentences, we are able to get better accuracies in comparison to other techniques.

Table 2: Comparative analysis of RankSum with state-of-the-art algorithms on the DUC 2002 dataset.

| Method | ROUGE-1 | ROUGE-2 | ROUGE-L |
| --- | --- | --- | --- |
| **LEAD** | 43.6 | 21.0 | 40.2 |
| **ILP** | 45.4 | 21.3 | 42.8 |
| **NN-SE** | 47.4 | 23.0 | - |
| **SummaRuNNer** | 47.4 | 24.0 | 14.7 |
| **Egraph+coh** | 47.9 | 23.8 | - |
| **Tgraph+coh** | 48.1 | 24.3 | - |
| **URANK** | 48.5 | 21.5 | - |
| **SummCoder** | 51.7 | 27.5 | 44.6 |
| **HSSAS** | 52.1 | 24.5 | 48.8 |
| **CoRank** | 52.6 | 25.8 | - |
| **Rank-emb** | 49.9 | 24.8 | 45.6 |
| **Rank-topic** | 51.4 | 25.9 | 47.2 |
| **Rank-keyword** | 52.0 | 26.3 | 48.6 |
| **RankSum** | **53.2** | **27.9** | **49.3** |

In Figure 2, we illustrate the ROUGE scores generated for 20 randomly se-lected documents from DUC 2002 dataset. The rouge scores are computed using the rank produced using the topic, keywords, embeddings, and RankSum algorithm. The graph depicts that the fusion of ranks of several features such as topic, keywords, embeddings, and position generate better ROUGE scores than any of the features individually. This shows that our RankSum algorithm is quite robust to capture the multiple aspects of words/sentences in the document, boosting the overall summarization accuracy of documents. It can be observed from the graph that individual features cannot attain good ROUGE-scores constantly and thus cannot produce a good compression of the query document. The fusion of various sentence ranking methods can yield better accuracy in most of the documents and thus retain the relevant information in a document generating better and more valuable summaries. We also present the summarization of a randomly selected DUC 2002 document using the RankSum algorithm in Table 3. As can be seen, that many of the phrases/sentences that appear in the Gold summary are similar to those of the ones generated using the RankSum framework.

Regarding the CNN/DailyMail dataset, we observed an improvement in accuracy with ROUGE-1, ROUGE-2, and ROUGE-L scores of 44.5, 24.0, and 41.0. As can be seen in Table 4, we obtained the best or comparable ROUGE scores with other state-of-the-art methods on CNN/DailyMail dataset. Our method ranked first for ROUGE-1, and ROUGE-2 scores whereas lags behind PEGASUSLARGE for ROUGE-L with a minimal margin of 0.1. It can be observed that our method is quite robust as it outperforms all the state-of-the-art supervised methods, including PACSUM, HIBERTM, HSSAS, BertSum, PEGASUSLARGE, JECS, and BART. This gives our framework an edge over existing summarization techniques. It does not require any ground truth data for producing extractive summaries. In contrast, all the other recently proposed methods require supervision with labeled data. Our summarization method is quite comprehensive as it explores the several significant aspects of sentences in a document required for summarization, such as topic, keywords, semantics and position. Also the individual rank extractors Rank-emb, Rank-topic, and Rank-keyword performed poorer as compared to the rank-based fusion technique.

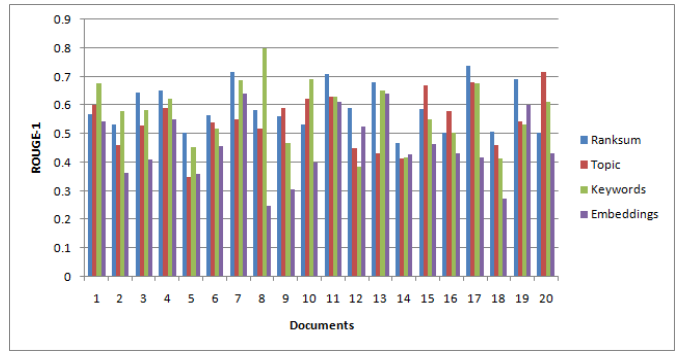

(a) ROUGE-1 graph

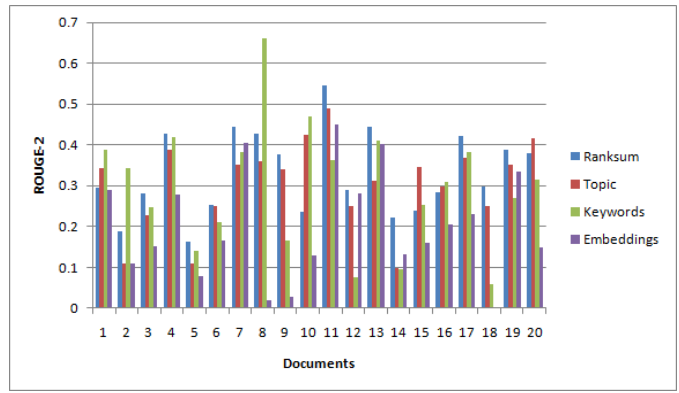

(b) ROUGE-2 graph

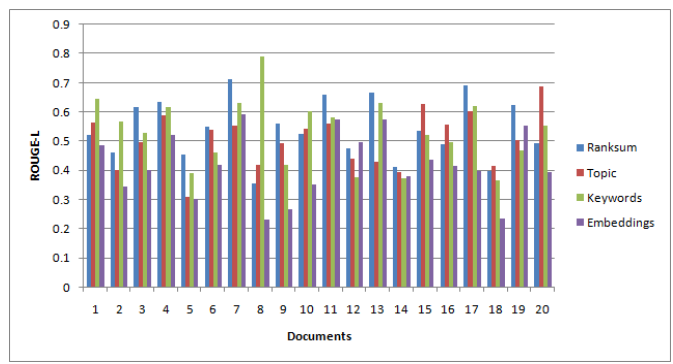

(c) ROUGE-L graph

Figure 2: Illustration of ROUGE-1, ROUGE-2 and ROUGE-L metrics considering ranking methodology based on topics, keywords, embeddings, and RankSum on 20 randomly selected documents of DUC 2002

Table 3: Gold summary and RankSum generated summary for a document from DUC 2002 dataset

| Gold Summary |
| --- |
| An overloaded ferry taking 183 Qiongzhong County students and teachers on a field trip to visit a hydroelectric power station capsized Wednesday, killing 55. The passengers and four crew members exceeded the ferry's capacity. The ship sank before it had sailed 200 yards, off Hainan Island in southern China. On September 22, another overloaded ferry sank in the Guangxi Zhuang Autonomous Region bordering Vietnam, killing 61. After 133 died July 21st, when a ferry capsized on the Min River in SW Sichuan Province, and 71 died July 25th, when a passenger boat sank on the Yangtze River, the Ministry of Communications began investigating river vessel safety. |
| RankSum Summary |
| On September 22, another overloaded ferry sank in the Guangxi Zhuang Autonomous Region bordering Vietnam, leaving 61 people dead and one missing. The accident occurred off Hainan island, the report said. It did not say how many people the boat was designed to hold. The move followed the July 21 capsizing of a ferry on the Min River in Southwestern Sichuan province in which 133 people died, and the July 25 sinking of a passenger boat on the Yangtze River in which 71 people drowned. The Ministry of Communications, which is responsible for inland water navigation, announced in August it had begun an investigation into the safety of China's river vessels. |

A graphical comparison of ROUGE scores obtained through the ranking based on individual parameters and their fusion using the RankSum methodology is depicted in Figure 3. The fusion of all the parameters using the RankSum framework is capable of increasing the accuracy on the randomly selected 20 documents and the whole testing CNN/DailyMail dataset. This shows that the weighted rank fusion of different aspects of documents can provide us with a broader and abstractive view of the document to produce better summaries than the individual features. The pertinent content encapsulated using one feature in a document is missed and can be captured via other parameters.

Thus, the fusion of several document features can better understand the syntax and semantics of the document to generate optimal extractive summaries. The RankSum summary of a randomly selected CNN/DailyMail document is shown in Table 5.

Table 4: Comparative analysis of RankSum with state-of-the-art algorithms on CNN/DailyMail

| Methods | ROUGE-1 | ROUGE-2 | ROUGE-L |
|---|---|---|---|
| **NN-SE** | 35.5 | 14.7 | 32.2 |
| **Bi-AES** | 38.8 | 12.6 | 33.85 |
| **LEAD** | 39.2 | 15.7 | 35.5 |
| **SummaRuNNer** | 39.6 | 16.2 | 35.3 |
| **REFRESH** | 40.0 | 18.2 | 36.6 |
| **PACSUM(BERT)** | 40.7 | 17.8 | 36.9 |
| **RNES w/o coherence** | 41.2 | 18.8 | 37.7 |
| **JECS** | 41.7 | 18.5 | 37.9 |
| **NeuSum** | 41.5 | 19.0 | 37.9 |
| **HSSAS** | 42.3 | 17.8 | 37.6 |
| **HIBERTM** | 42.3 | 19.9 | 38.8 |
| **BertSum** | 43.2 | 20.2 | 39.6 |
| **BERTSUM+Classifier** | 43.2 | 20.2 | 39.6 |
| **BART** | 44.1 | 21.2 | 40.9 |
| **PEGASUSLARGE** | 44.1 | 21.4 | **41.1** |
| **MATCHSUM** | 44.4 | 20.8 | 40.5 |
| **Rank-emb** | 42.2 | 21.5 | 40.3 |
| **Rank-topic** | 43.4 | 22.9 | 40.2 |
| **Rank-keyword** | 43.7 | 23.7 | 40.7 |
| **RankSum** | **44.5** | **24.0** | 41.0 |

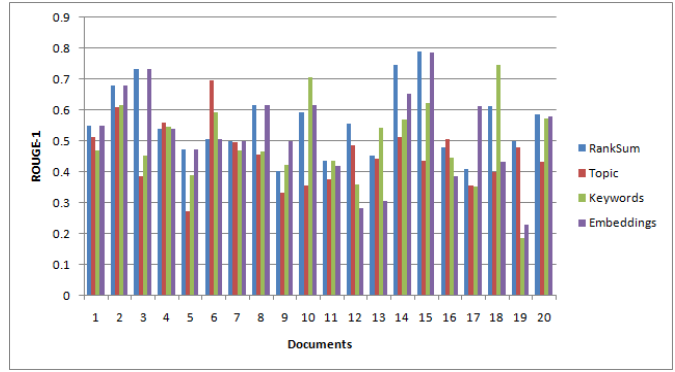

(a) ROUGE-1 graph

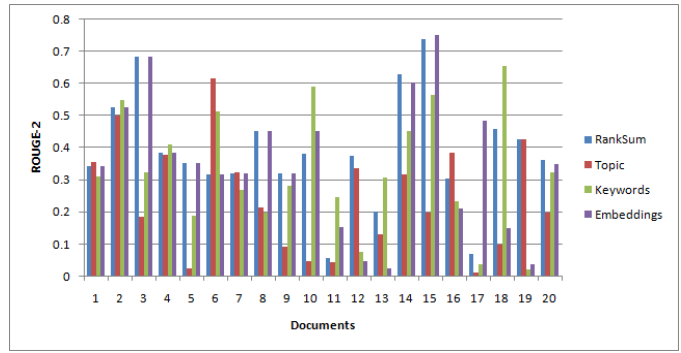

(b) ROUGE-2 graph

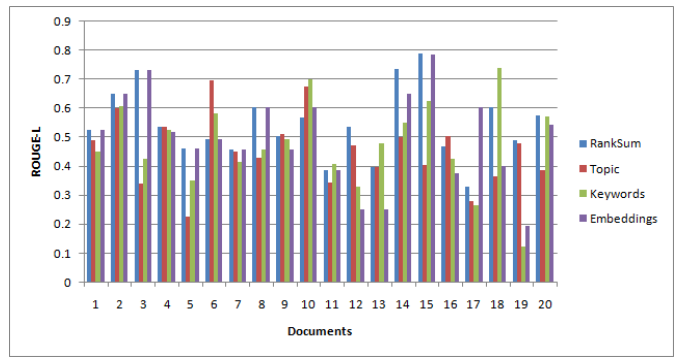

(c) ROUGE-L graph

Figure 3: Illustration of ROUGE-1, ROUGE-2, and ROUGE-L metrics considering RankSum, topic, semantics and keyword approach for the ranking sentences on 20 randomly selected documents of CNN/DailyMail

Table 5: Gold summary and RankSum generated summary for a document from CNN/DailyMail dataset

| Gold Summary |
| --- |
| Astonishing images have emerged of hollywood legend mickey rooney having a gash to his head stitched amid allegations he suffered elder abuse before his death. the actor who died earlier this year at the age of is shown in one picture having a large wound on his head treated by medics. In a second image taken in january mickey rooney is pictured with a missing tooth and other facial injuries. The shocking photos were revealed for the first time amid claims the star may have suffered abuse in the years before he died in april this year. |
| RankSum Summary |
| In the enquirers article rooneys eighth wife jan chamberlin vehemently denies any suggestion that she may have abused the star. In a second image he is pictured with a missing tooth and other facial injuries. According to the national enquirer some members of the stars family are preparing to hand a file over to law enforcement chiefs which they believe may explain his death. Astonishing images have emerged of hollywood legend mickey rooney having a gash to his head stitched amid allegations he suffered elder abuse before his death. In a second image taken in january mickey rooney is pictured with a missing tooth and other facial injuries. |

## 5. Conclusions

In this paper, we presented RankSum, a unified and unsupervised framework for extractive text summarization. Bearing in mind that humans combine different characteristics for text summarization tasks, RankSum is based on combining several structural and semantic features of a document. Our proposal captures multi-dimensional information from the document using keywords, signature topics, sentence embeddings, and a sentence's position in the document. All of the features individually are capable of extracting important content from the document. However, when combined through a rank fusion scheme, they can cover different aspects of a document to summarize it adequately.

We designed a ranking method of sentences for summarization based on topic vectors estimated using probabilistic topic vectors. A novel method for ranking sentences based on sentence embeddings computed through Siamese networks has also been introduced. The keyword information derived, based on document graph representation, has also been exploited to rank the summary generation sentences. We also developed a novel scheme to eliminate redundant sentences by using bigrams, trigrams and sentence embeddings.

Experimentally, it has been shown that the RankSum method yields a more robust description and outperforms most of the existing state-of-the-art approaches. It was also illustrated that different ranking strategies formulated via topic, keywords, and embeddings may not capture the important content in the document, individually, in most cases. However, a combination and fusion of all the presented ranking strategies can deliver better ROUGE scores comparatively and, thus, can be a beneficial addition for summariza-tion. Another benefit of our approach is that it does not require labeled data for training.

In the future, we will explore the applicability of the RankSum framework for abstractive summarization.

**Acknowledgements**

This research is supported by the framework agreement between the University of León and INCIBE (Spanish National Cybersecurity Institute) under Addendum 01.